%% file: main.tex
\documentclass[10pt,twocolumn,letterpaper]{article}

\usepackage{cvpr}              

\input{preamble}
\definecolor{cvprblue}{rgb}{0.21,0.49,0.74}
\usepackage[pagebackref,breaklinks,colorlinks,allcolors=cvprblue]{hyperref}

\def\paperID{*****} 
\def\confName{CVPR}
\def\confYear{2026}

\title{Monte Carlo Pass Search: \\
Using Trajectory Generation for 3D Counterfactual Pass Evaluation in Football}

\author{Andrew Kang\\
Carnegie Mellon University\\
Pittsburgh, USA\\
{\tt\small akang2@andrew.cmu.edu}
\and
Priya Narasimhan\\
Carnegie Mellon University\\
Pittsburgh, USA\\
{\tt\small priyan@andrew.cmu.edu}
}

\begin{document}
\maketitle
\input{sec/0_abstract}    
\input{sec/1_intro}
\input{sec/2_relatedWork}

\input{sec/3_mcps}

\input{sec/4_results}

\input{sec/5_discussion}
{
    \small
    \bibliographystyle{ieeenat_fullname}
    \bibliography{main}
}


\end{document}

%% file: preamble.tex
\usepackage{bbm}

%% file: sec/0_abstract.tex
\begin{abstract}
We recast pass evaluation in football (soccer) as a Monte Carlo Tree Search (MCTS)-like evaluation problem whose components mostly exist in the literature under different names: a value model (possession value), a world model (multi-agent trajectories with ball interactions), and a policy over counterfactual actions (sampling pass variants with noise). Building on the first public high-fidelity tracking dataset with 3D ball trajectories from the Bundesliga, we introduce Monte Carlo Pass Search (MCPS), which infers kick parameters for each observed pass, samples execution variants and option variants, rolls each candidate forward with a ball-conditioned world model until the next ball interaction, and scores outcomes with a learned value model to obtain a distribution over gained value. This distribution enables distribution-aware attribution with two complementary execution-surplus scores used for analysis and ranking: mean-based and percentile-based scores. To make the world model sample-efficient under limited public data, we adapt a discrete-token, autoregressive trajectory generator from autonomous driving (SMART) and show it yields strong best-of-20 forecasting accuracy compared to baselines, while supporting fully hypothetical rollouts for downstream evaluation. We have released model checkpoints and code in 
\href{https://github.com/andrewkang12345/monteCarloPassSearch}{this repository}.
\end{abstract}

%% file: sec/1_intro.tex
\section{Introduction}
\label{sec:intro}

\begin{figure*}
  \centering
  \includegraphics[width=0.8\linewidth]{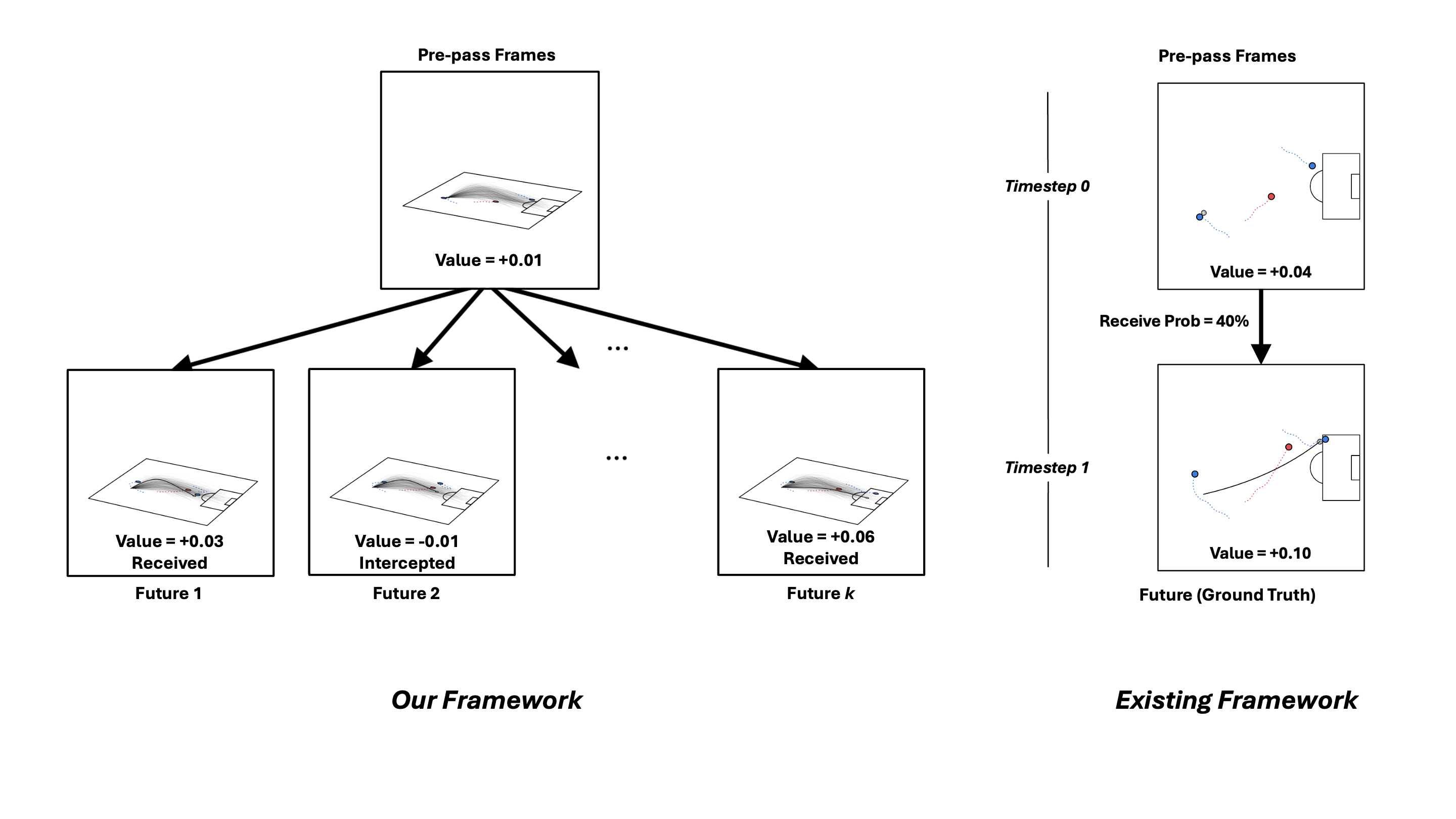}
  \caption{\textbf{From point estimates to distributions.} Existing pass-evaluation workflows (right) primarily score the observed outcome, with limited counterfactual reasoning and little notion of execution sensitivity. We introduce Monte Carlo Pass Search (MCPS; left): for each observed pass we infer kick parameters, sample 256 counterfactual executions and alternatives, generate short-horizon futures with a trajectory-conditioned world model, and evaluate the induced value distribution.}
  \label{fig:fig1}
\end{figure*}

Computer vision (CV) with multi-camera systems has made spatiotemporal tracking data a standard input for professional football analysis across coaching, scouting, and recruitment.
Yet, unlike many high-impact CV domains, football analytics remains constrained by a proprietary ecosystem: datasets, model checkpoints, and evaluation protocols are rarely released, limiting reproducibility and slowing progress.
The recent public release of higher-fidelity tracking data \cite{bassek2025integrated} creates an unusually timely opportunity to reconnect football analytics with the broader CV community and to establish baselines that are comparable and extensible.

\paragraph{Why passing is hard to evaluate.}
Passing decisions are confounded by both execution noise and downstream interaction.
Two visually similar passes can differ dramatically in how narrow their ``success window'' is; conversely, a technically excellent attempt can appear poor if a teammate loses the subsequent duel.
This outcome bias makes it difficult to separate (i) whether the passer chose a good option, (ii) whether the chosen option was intrinsically fragile, and (iii) whether the execution was better or worse than typical for that attempt.

\paragraph{What existing models provide and miss.}
Widely used metrics such as expected threat (xT) \cite{rudd2011framework, singh2019expected, bransen2020valuing} and expected pass (xPass) \cite{anzer2022expected} are valuable but are typically deployed as point estimates: they either score the realized action or assign a single expected outcome to a hypothetical pass.
A notable exception is the physics-based pass probability model of Spearman et al.\ \cite{spearman2017physics}, which treats a pass as a probabilistic competition among nearby players to intercept and control the ball, enabling hypothetical passing analysis and perturbation-based difficulty estimates.
However, two gaps remain for such existing distributional pass evaluation.
First, ``ideal pass'' analysis can obscure the fact that there may be multiple distinct high-quality executions of the same option, each with different robustness.
Second, rule-based motion assumptions (e.g., simplified effort/reaction models) limit how well counterfactual futures capture game-contextual off-ball movement and downstream value.

\paragraph{Our perspective.}
We treat a pass as a \emph{distribution} over plausible executions and plausible short-term futures, rather than a single realized event.
To do so, we borrow the high-level idea of search-and-evaluate from Monte Carlo Tree Search (MCTS), while replacing hand-designed simulators with learned, trajectory-conditioned generative models.
Unlike MuZero-style planning in a purely latent space, our world model operates in raw spatiotemporal space: given a pre-pass context and a candidate ball flight, it generates stochastic multi-agent futures until the next ball interaction.
Evaluating these rollouts with a learned possession-value function yields a value distribution that makes risk and robustness explicit.

\paragraph{MCPS in one sentence.}
From a pre-pass state, MCPS samples counterfactual pass executions and counterfactual alternatives, rolls them out with a trajectory-conditioned world model, and scores them with a value model to obtain a distribution over the gained possession value, $\Delta\mathrm{PV}$, for attribution and comparison.

\paragraph{Contributions.}
\begin{itemize}[leftmargin=*,itemsep=2pt,topsep=2pt]
  \item \textbf{Trajectory-conditioned rollout evaluation for passing:} a world-model-based framework that generates diverse short-horizon futures conditioned on candidate ball trajectories.
  \item \textbf{Dual Monte Carlo search:} local sampling around an observed pass to model execution sensitivity, and global sampling to approximate the opportunity set of alternatives.
  \item \textbf{Distribution-aware pass evaluation:} two execution-surplus metrics (mean-difference and percentile) and practical visualizations (opportunity/sensitivity views) that support coaching and recruitment workflows.
\end{itemize}

%% file: sec/2_relatedWork.tex
\section{Related Work}
\label{sec:relatedWork}

\paragraph{Football analytics from tracking data.}
Our Monte Carlo Pass Search (MCPS) framework sits at the intersection of (i) pass success / reception modeling, (ii) state/action value modeling, and (iii) generative forecasting of spatiotemporal trajectories. Much of the classic football analytics literature relies on analytically-structured models or point-estimate predictors, and many strong modern systems are trained on proprietary data and are not released publicly.

\textit{Pass probability and pitch control.}
Pass probability models estimate the likelihood that a pass is completed and, in some cases, which player will receive or intercept the ball \cite{spearman2017physics,anzer2022expected,goes2019not}.
Among these, Spearman et al.\ \cite{spearman2017physics} is particularly foundational: they frame passing as a probabilistic competition between nearby players to \emph{intercept} and then \emph{control} the ball, yielding an interpretable, predictive model that can be queried at kick time for both observed and \emph{hypothetical} passes.
Their formulation also enables spatial pitch-control fields and downstream passing metrics, and it directly motivates two ingredients we adopt and extend: (1) evaluating hypothetical executions of a pass by perturbing ball-flight parameters, and (2) comparing an observed decision to a best-available alternative under the same pre-pass state.

\textit{Value models for states and actions.}
A complementary line of work assigns scalar values to states, possessions, or actions to quantify contribution to scoring chances or match outcomes \cite{fernandez2021framework,rahimian2021inferring,bransen2020valuing,decroos2019actions}.
These value functions are often used as the terminal or immediate reward signal for decision evaluation (e.g., expected threat / possession value), but they are typically applied as point estimates to realized events rather than as distributions over counterfactual futures.

\textit{Generative trajectory forecasting.}
Generative trajectory models predict short-horizon futures of player (and sometimes ball) motion from spatiotemporal context \cite{omidshafiei2021time,capellera2024transportmer,xu2024sports}.
However, strong learned forecasting models are often not publicly released, and training data is frequently proprietary; an exception is \cite{xu2024sports}.
In the absence of accessible tracking benchmarks, many works instead model discrete on-ball events (e.g., passes, shots) without generating full continuous trajectories \cite{oh2015graphical,wei2015forecasting,rahimian2021inferring}.
This landscape is beginning to change with the release of higher-fidelity spatiotemporal tracking datasets such as \cite{bassek2025integrated}, which we leverage in this paper.

\textit{Counterfactual and ensemble evaluation frameworks.}
Several frameworks combine components like pass success models, value functions, and/or forecast models to evaluate counterfactual (``what-if'') futures \cite{teranishi2022evaluation,fujii2023adaptive,wang2024tacticai,spearman2017physics}.
Spearman et al.\ \cite{spearman2017physics} already demonstrates a sophisticated hypothetical-passing workflow, including searching for an ``ideal'' execution and probing stability via perturbations, but relies on hand-specified assumptions about player response (e.g., full-effort interception/control) and primarily grounded-pass physics.
MCPS generalizes this style of analysis by (i) evaluating \emph{downstream value} via simulated rollouts rather than only pass success, and (ii) producing \emph{distributions} over outcomes (mean and tail risk) under both local execution noise and global alternative-option search.
Teranishi et al.\ \cite{teranishi2022evaluation} similarly motivates the use of learned generative models for evaluation, but focuses on off-ball movement assessment and typically summarizes futures with a single ``average'' prediction; in contrast, MCPS explicitly samples many plausible futures per candidate pass to estimate risk and fragility.

\paragraph{Generative trajectory modeling in autonomous driving.}
Compared to sports, autonomous driving offers mature public benchmarks and standardized evaluation for forecasting \cite{chang2019argoverse,wilson2023argoverse,caesar2020nuscenes,sun2020scalability}.
Two dominant modeling paradigms are (i) continuous diffusion-based one-shot prediction and (ii) autoregressive discrete-token prediction.
We adopt the autoregressive perspective to better enforce causal structure around the ball trajectory and to improve sample-efficiency under limited training data, adapting the architecture of a recent high-performing autoregressive model (SMART \cite{wu2024smart}) to our setting.

\begin{figure}
  \centering
  \includegraphics[width=0.8\linewidth]{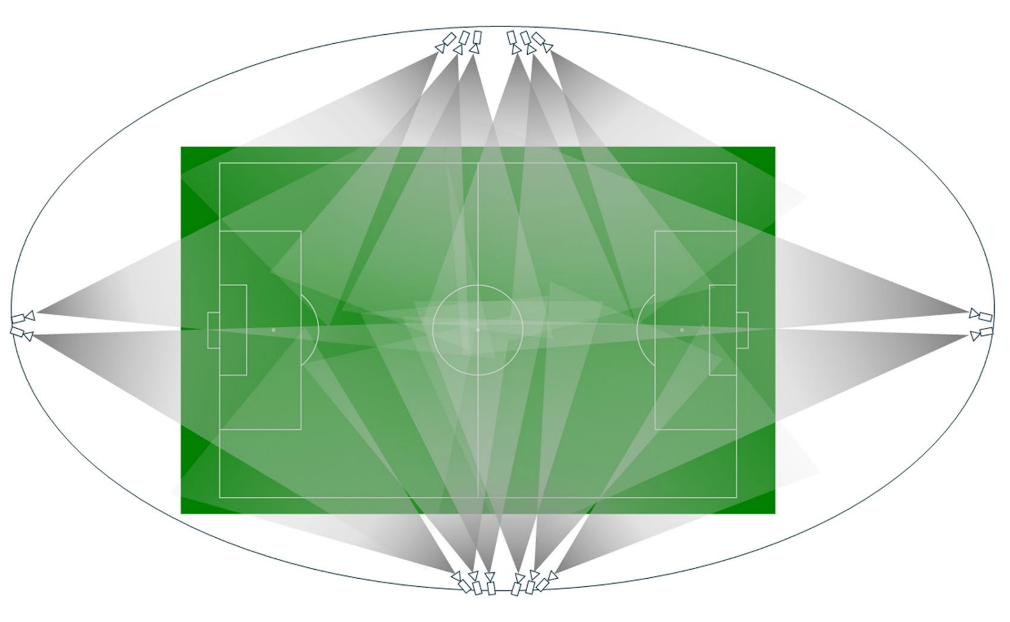}
  \caption{TRACAB Gen5 camera setup \cite{linke2020football} used to estimate player coordinates and 3D ball coordinates in the public tracking dataset \cite{bassek2025integrated} used in this paper.}
  \label{fig:fig2}
\end{figure}

\paragraph{Monte Carlo Tree Search, model-based RL, and world models.}
Our planning perspective is inspired by policy learning methods, although we use it primarily for offline action evaluation. MuZero, which combines tree search with a learned dynamics model to attain a strong policy across Atari and board games \cite{schrittwieser2020mastering}.
MuZero learns latent dynamics and produces policy/value predictions suitable for search.
In contrast, we do not plan purely in a latent space: our generator predicts future \emph{raw} player and ball trajectories conditioned on candidate ball flights, and we explicitly integrate an external football value model into the Monte Carlo evaluation loop.

More broadly, our work relates to world-model-based reinforcement learning methods that learn generative environment dynamics and use imagination for decision-making.
DreamerV3 \cite{hafner2023mastering} demonstrates broad task generality via latent dynamics and imagined rollouts.
DIAMOND \cite{alonso2024diffusion} highlights how high-fidelity generative modeling (via diffusion) can improve downstream policy performance.
These results motivate our use of learned dynamics for planning, while our focus differs: we use short-horizon, trajectory-conditioned rollouts to evaluate recorded passing decisions and separate decision quality from execution quality.

\paragraph{Physics-based strategy and simulation in other sports.}
Using physics models and simulators to inform strategy appears across sports.
Fragkiadaki et al.\ learn predictive physics models in billiards and plan via ``visual imagination'' \cite{fragkiadaki2015learning}.
In ten-pin bowling, Ji et al.\ derive a rigid-body simulator to map release conditions to strike probability and analyze robustness (``miss room'') \cite{ji2022using}.
In curling, Xiao et al.\ combine perception with a domain-specific MCTS planner (inspired by Yee et al.\ \cite{yee2016monte}) in continuous action spaces and report competitive effectiveness \cite{xiao2023policy}, alongside other digital and physical approaches \cite{han2022game,lee2018deep,doi:10.1126/scirobotics.abb9764}.
Our work shares the goal of robust action selection, but emphasizes learned, data-driven dynamics (rather than fixed simulators) and distributional evaluation of pass outcomes (risk and tail behavior) within a unified planning-and-valuation loop.

%% file: sec/3_mcps.tex
\begin{figure}
  \centering
  \includegraphics[width=0.8\linewidth]{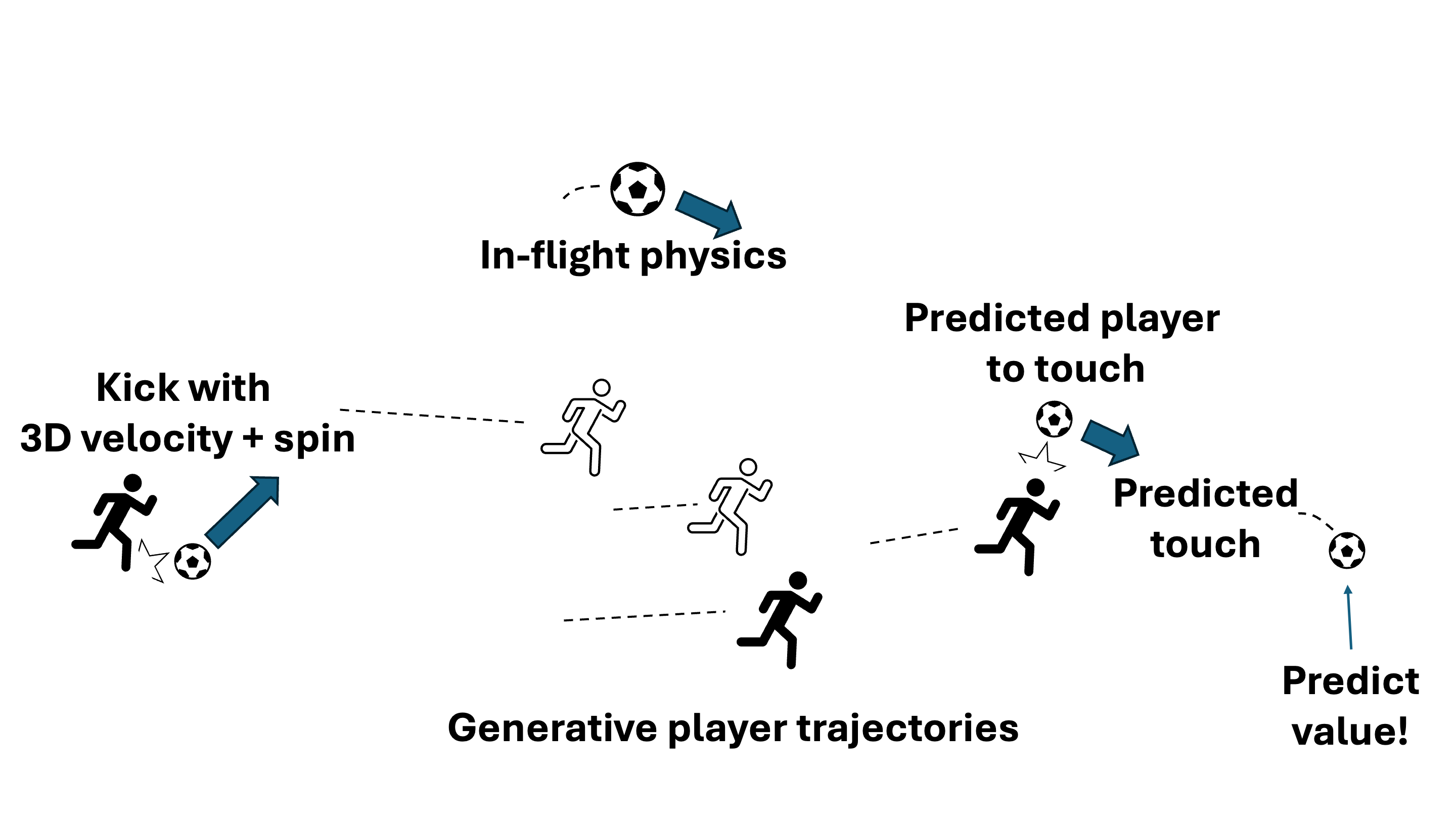}
  \caption{Workflow for each simulated pass in our Monte Carlo Pass Search (MCPS) framework.}
  \label{fig:fig3}
\end{figure}

\section{Monte Carlo Pass Search}
\label{sec:mcps}

MCPS is inspired by world-model-based planning in that it evaluates candidate actions by simulating futures with a dynamics model and scoring them with a value function.
Unlike standard planning, which aims to do \emph{one-shot prediction} of an optimal action, MCPS is designed to \emph{evaluate an observed pass} by comparing it to distributions of counterfactual executions (``same option, different kick'') and counterfactual alternatives (``different option'').

\paragraph{Dataset.}
We use the public tracking dataset of \cite{bassek2025integrated}, which provides official match metadata, event annotations, and 25\,Hz positional tracking for all players and the ball, including 3D ball coordinates, across seven matches from the German Bundesliga's first and second divisions.
For each pass, we extract a fixed pre-pass context window of 64 frames.
Because event timestamps may not coincide with the true kick frame in tracking, we adjust the kick index with a simple heuristic based on ball kinematics around the event time. Our train/val/test splits are 5/1/1 matches.

\subsection{The Value Model}
\label{subsec:valueModel}

The value model assigns a scalar score to a \emph{state} of play given sufficient context.
In football, we define state value as the ego team's short-horizon attacking opportunity minus the opponent's.
We operationalize this using a widely used definition of \emph{Possession Value} (PV): the probability that the ego team scores within 10 seconds minus the probability that the opponent scores within 10 seconds.

\paragraph{Implementation.}
We train a Transformer on tracking windows with two binary cross-entropy heads (home and away), using shot-within-10s labels weighted by an expected-goals proxy (shots serve as fractional goals to mitigate sparse scoring).
We apply random horizontal and vertical flips to all player and ball coordinates during training for data augmentation \cite{wang2024tacticai}.

\subsection{The World Model}
\label{subsec:worldModel}

The world model predicts the short-term evolution of the scene under a candidate pass, enabling counterfactual rollouts.
Because the stochasticity of raw futures in football is already challenging (due to human motion and contested touches), we model transitions directly in \emph{raw} space (player and ball trajectories), rather than only in a latent space.

We treat a pass as a single transition step that ends at the first meaningful interaction (a controlled touch, clear deflection, or the ball leaving play).
This requires modeling both (i) \emph{who} touches next and \emph{when}, and (ii) \emph{how} the ball changes at contact.
Accordingly, our simulation stack includes: a player trajectory generator, a \textbf{Player-to-Touch} predictor, and a \textbf{Ball-at-Touch} predictor.

\paragraph{Implementation.}
For trajectory generation, we adapt SMART \cite{wu2024smart} to football tracking and use an autoregressive discrete-token formulation.
We downsample to 12.5\,Hz and define one token step as 5 frames (0.4\,s).
Player motion tokens are 10-dimensional (planar velocities over 5 frames) with a 2048-code vocabulary; ball motion tokens are 15-dimensional (3D velocities over 5 frames) with a 1024-code vocabulary.
Vocabularies are built via $k$-means on standardized displacements.
A decoder-only space--time Transformer conditions on 8 history tokens and predicts 24 rollout tokens under teacher forcing, trained with masked cross-entropy with embeddings for entity type (player/ball) and team indicator.

The \textbf{Player-to-Touch} module predicts the next toucher (and time-to-touch) from per-frame touch supervision constructed via a velocity-change + proximity heuristic; we train it with a survival-style objective (BCE on touch hazard plus weighted CE over toucher identity on touch frames).
The \textbf{Ball-at-Touch} module predicts the post-touch ball velocity with a diagonal-Gaussian regressor trained by masked Gaussian negative log-likelihood.

\subsection{The Policy Model (Search over Pass Variants)}
\label{subsec:policyModel}

MCPS uses Monte Carlo search to construct counterfactual action sets for evaluating an observed pass.
For each observed pass, we perform (i) a \textbf{local} search that perturbs the observed kick parameters to model execution noise, and (ii) a \textbf{global} search with larger perturbations to explore alternative options from the same pre-pass state.

A prerequisite is accurate inference of the kick parameters (initial 3D velocity and a spin proxy).
Directly using observed ball velocity at the kick frame is unreliable, and simple physics does not match the measured trajectory.
We therefore fit kick parameters with a CEM-style solver coupled to a ball-flight simulator (with global constants such as gravity, drag, restitution, and friction) by minimizing trajectory error up to the first interaction.
We retain only passes whose fitted trajectory matches the observed ball flight with near-zero error.

\paragraph{Implementation.}
For each retained pass, we sample 256 local variants and 256 global variants (plus the fitted observed pass).
Local variants apply small perturbations to direction, speed, vertical component, and spin.
Global variants apply wide direction changes with moderate speed spread and small spin spread.
We enforce hard caps on sampled parameters based on high-percentile statistics of the fitted pass manifold (e.g., planar speed, vertical speed, and spin magnitude).

\subsection{Evaluation Framework}
\label{subsec:evalFramework}

Let $s$ denote the pre-pass state (the 64-frame context window), and let $\theta$ denote the inferred kick parameters for a pass execution.
For a given $\theta$, we simulate the pass forward with our world model until the first meaningful ball interaction (touch/deflection/out-of-bounds), producing a next-state $s'(\theta)$.
We then score the transition using our Possession Value (PV) model as \emph{PV added}:
\begin{equation}
\Delta \mathrm{PV}(\theta) \;=\; \mathrm{PV}\!\left(s'(\theta)\right) - \mathrm{PV}(s).
\end{equation}
The observed pass is evaluated by comparing its $\Delta \mathrm{PV}$ to the empirical distribution induced by Monte Carlo variants.

\paragraph{Local and global counterfactual sets.}
For each observed pass, we generate two sets of counterfactual executions:
(i) a \textbf{local} set $\{\theta^{(i)}_{\text{loc}}\}_{i=1}^{N}$ obtained by small perturbations around the fitted observed parameters (intended to represent execution noise), and
(ii) a \textbf{global} set $\{\theta^{(j)}_{\text{glob}}\}_{j=1}^{M}$ obtained by larger perturbations (intended to represent alternative options).
We compute $\Delta \mathrm{PV}$ for each variant by running the same rollout-and-score procedure.

\paragraph{Per-pass execution surplus.}
We use two complementary, distribution-aware scores for a single observed pass, both computed relative to a chosen counterfactual set (local or global).
Let $\theta_{\text{obs}}$ denote the fitted observed execution and let $\Delta \mathrm{PV}_{\text{obs}}=\Delta \mathrm{PV}(\theta_{\text{obs}})$.

\textbf{(i) Mean-difference surplus.}
We define the observed-vs-mean differential:
\begin{equation}
S_{\text{mean}} \;=\; \Delta \mathrm{PV}_{\text{obs}}
\;-\;
\frac{1}{K}\sum_{k=1}^{K} \Delta \mathrm{PV}\!\left(\theta^{(k)}\right),
\end{equation}
where $\{\theta^{(k)}\}_{k=1}^{K}$ denotes either the local set or the global set.
This matches our first ranking metric in Sec.~\ref{sec:results}: the average differential between observed pass PV and mean counterfactual pass PV.

\textbf{(ii) Percentile surplus.}
We also compute the percentile rank of the observed pass value within its counterfactual distribution:
\begin{equation}
S_{\text{pct}} \;=\;
\frac{1}{K}\sum_{k=1}^{K}
\mathbbm{1}\!\left[\Delta \mathrm{PV}_{\text{obs}} \ge \Delta \mathrm{PV}\!\left(\theta^{(k)}\right)\right],
\end{equation}
which lies in $[0,1]$ (higher is better).
This matches our second ranking metric in Sec.~\ref{sec:results}: the average percentile of a player's observed pass PV among its counterfactual variants.

\paragraph{Player-level aggregation.}
To rank passers in a match, we aggregate per-pass scores across all passes attempted by a player and report the mean $S_{\text{mean}}$ and mean $S_{\text{pct}}$ separately for local search and global search (Fig.~\ref{fig:global}, Fig.~\ref{fig:local2}).
Intuitively, local aggregation emphasizes execution relative to the fragility of a given intended pass, while global aggregation emphasizes how the observed choice compares to the opportunity set explored by global variants.

%% file: sec/4_results.tex
\section{Results}
\label{sec:results}

\begin{figure*}
  \centering
  \includegraphics[width=0.8\linewidth]{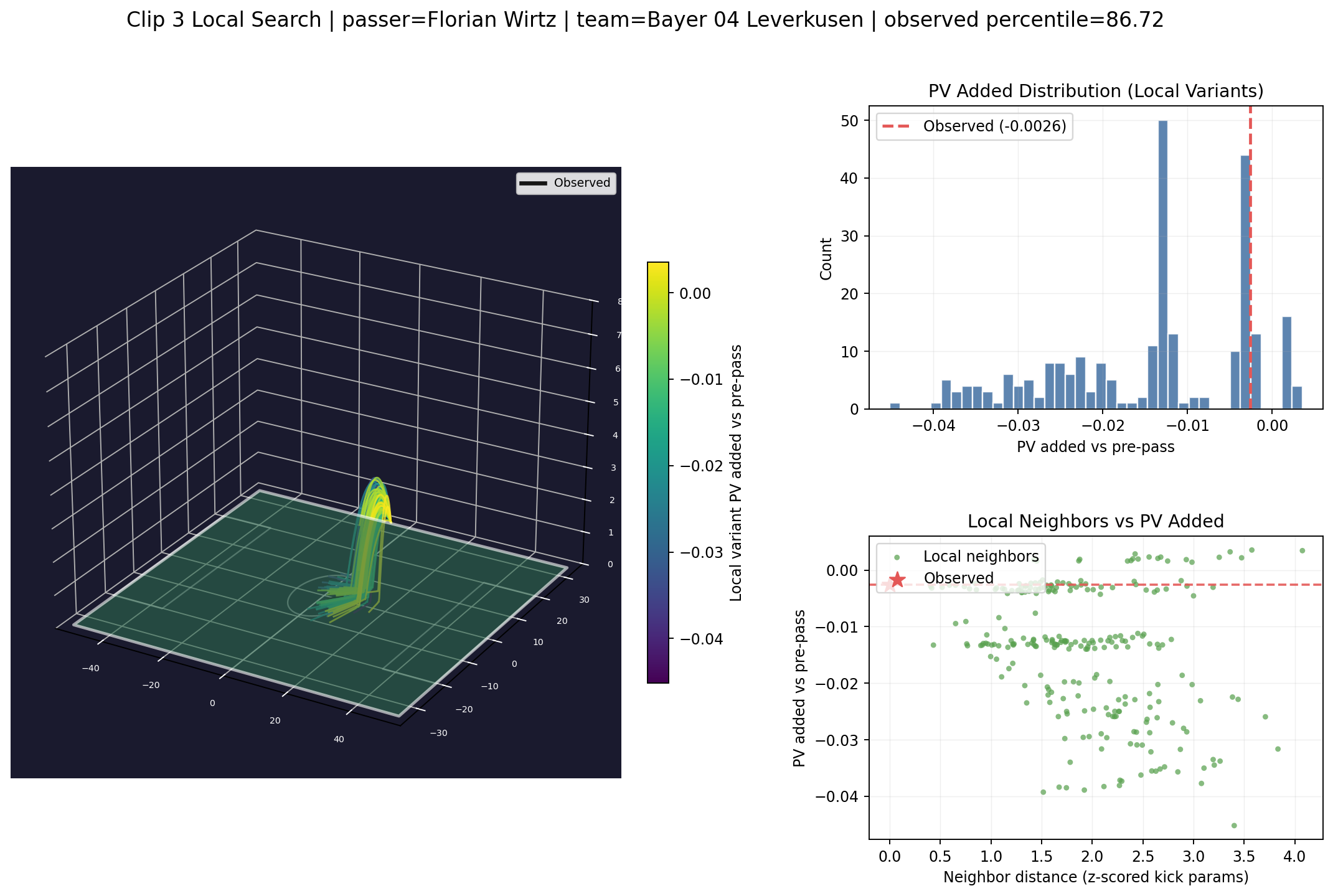}
  \caption{\textbf{Example MCPS evaluation (local search).} Left: observed ball trajectory and sampled local-variant trajectories (colored according to PV obtained, $\Delta\mathrm{PV}$). Top-right: distribution of counterfactual pass $\Delta\mathrm{PV}$ under local perturbations, with the observed pass shown as a dashed red line. Bottom-right: $\Delta\mathrm{PV}$ versus distance from the observed pass in inferred kick-parameter space, illustrating sensitivity of downstream value to small execution changes.}
  \label{fig:clip3}
\end{figure*}

\subsection{Case Study: Bochum vs.\ Leverkusen (Bundesliga 2022/23)}
We showcase MCPS on a held-out test match from \cite{bassek2025integrated}: VfL Bochum vs.\ Bayer Leverkusen (May 27, 2023), which Bochum won 3--0.
Goals were scored by St\"oger, Asano, and F\"orster, assisted by Zoller, St\"oger, and Asano.

\paragraph{Example pass analysis.}
Fig.~\ref{fig:clip3} illustrates a lofted observed pass by Florian Wirtz.
Even within a tight neighborhood of kick-parameter perturbations (local search), MCPS yields a non-trivial spread of counterfactual $\Delta\mathrm{PV}$ values, highlighting execution sensitivity under the learned dynamics and interaction models.

\paragraph{Player rankings.}
We rank passers in this match in Figures \ref{fig:global} and \ref{fig:local2} by aggregating the two distribution-aware metrics from Sec.~\ref{subsec:evalFramework} across all passes attempted by each player, and we report rankings separately for global and local search.
Concretely, we aggregate:
(i) \textbf{mean-difference surplus} (observed $\Delta\mathrm{PV}$ minus mean counterfactual $\Delta\mathrm{PV}$), and
(ii) \textbf{percentile surplus} (observed $\Delta\mathrm{PV}$ percentile among counterfactual variants).
Other risk-sensitive summaries (e.g., VaR/CVaR) are left to future work.

\paragraph{Observations.}
Across this match, MCPS highlights (i) passes whose value is robust to local execution noise (high local percentile surplus), and (ii) situations where global counterfactuals suggest missed higher-value alternatives (high opportunity gaps under global search). However, for passes that do not have accurate inferred parameters, evaluation tends to be noisy. Thus, we constrained our analysis to 512 passes that we could infer accurately.

\subsection{Baseline Comparison and Ablation}
\label{subsec:ablations}

Comparing football sub-models against the broader literature is challenging due to scarce public benchmarks and unreleased checkpoints (Sec.~\ref{sec:intro}).
We therefore report comparisons to (a) publicly available checkpoints where possible and (b) standard ablations.

\paragraph{Player trajectories.}
Most football trajectory models are trained on proprietary tracking and do not release checkpoints.
To the best of our knowledge, Sports-Traj \cite{xu2024sports} is the only public multi-agent checkpoint suitable for comparison.
Sports-Traj is trained primarily for short-horizon imputation/prediction; to evaluate longer horizons we roll it out autoregressively, which differs from its training objective.
We also include common ablations: static, constant velocity, and a naïve Transformer regressor.
Our model is a discrete-token autoregressive generator inspired by SMART \cite{wu2024smart}.

\begin{table}[t]
  \centering
  \small
  \setlength{\tabcolsep}{5pt}
  \begin{tabular}{l cc}
    \toprule
    Model & minADE$_{20}$ $\downarrow$ & minFDE$_{20}$ $\downarrow$ \\
    \midrule
    Static & 6.8 & 13.4 \\
    Constant velocity & 5.0 & 10.1 \\
    Na\"ive Transformer & 3.8 & 7.5 \\
    Sports-Traj \cite{xu2024sports} & 4.2 & 6.9 \\
    Ours (SMART-based) \cite{wu2024smart} & \textbf{2.4} & \textbf{4.7} \\
    \bottomrule
  \end{tabular}
  \caption{\textbf{Trajectory forecasting ablation.} Best-of-20 metrics on the test split (lower is better). A qualitative example of our model is in Figure~\ref{fig:fig5}.}
  \label{tab:traj_ablation}
\end{table}

\begin{figure*}
  \centering
  \includegraphics[width=0.8\linewidth]{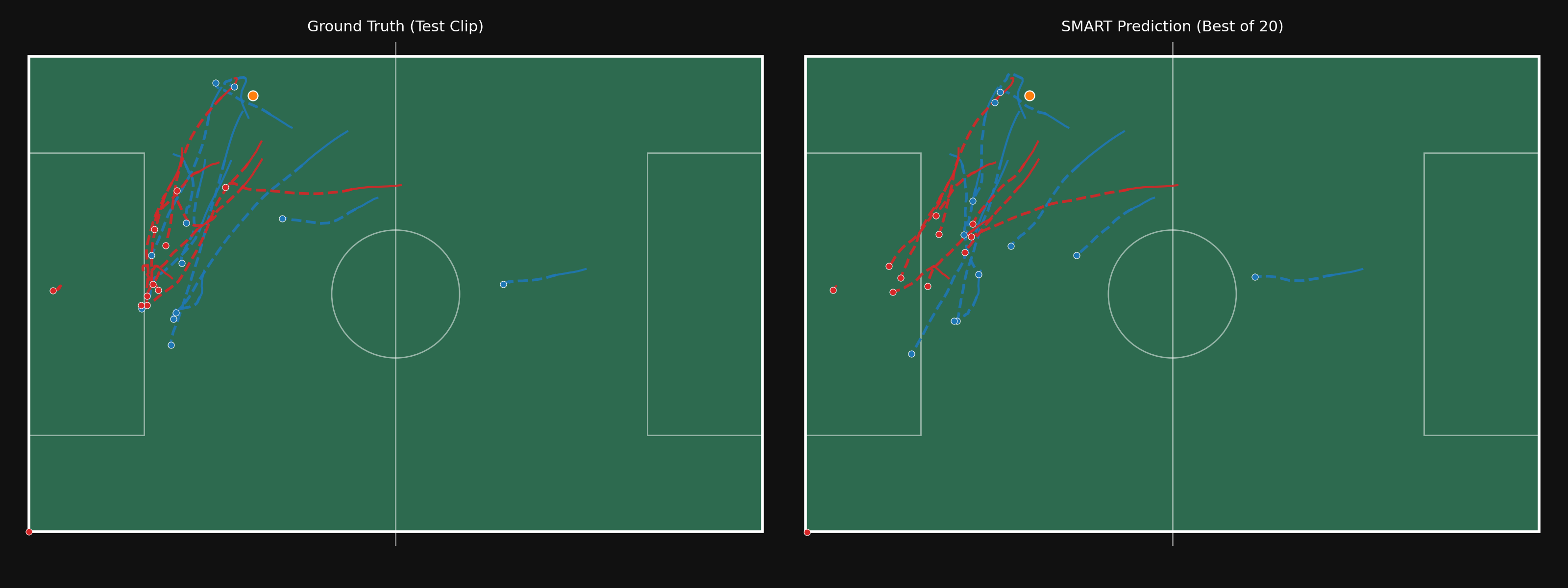}
  \caption{\textbf{Qualitative multi-agent trajectory forecasting.} This is an example clip generated with the ball trajectory as condition for each timestep. Left: ground truth. Right: best-of-20 sampled rollouts under ADE/FDE. Blue/red denote teams; yellow denotes the ball (conditioning).}
  \label{fig:fig5}
\end{figure*}

\paragraph{Player-to-Touch.}
Predicting which player touches the ball next (and when) has limited public baselines beyond the physics-based approach of Spearman et al.\ \cite{spearman2017physics}.
We reuse their reported numbers as a reference and compare to an Expected Pass-style formulation \cite{anzer2022expected} where applicable under available annotations, alongside our learned Player-to-Touch model.

\begin{table}[t]
  \centering
  \scriptsize
  \setlength{\tabcolsep}{3.5pt}
  \begin{tabular}{l c c c}
    \toprule
    \textbf{Model}
    & \shortstack{\textbf{Receiver}\\\textbf{Top-1 Acc.} $\uparrow$}
    & \shortstack{\textbf{Pass Success}\\\textbf{Acc.} $\uparrow$}
    & \shortstack{\textbf{Pass Success}\\\textbf{AUROC} $\uparrow$} \\
    \midrule
    Spearman, reported \cite{spearman2017physics}
    & 0.679
    & 0.805
    & $\sim$0.85 \\
    Anzer, reported \cite{anzer2022expected}
    & \textbf{0.899}
    & \textbf{0.915}
    & \textbf{0.934} \\
    Ours on Generated Futures\\
    (Player-to-Touch + Traj)
    & 0.605
    & 0.777
    & 0.799 \\
    \bottomrule
  \end{tabular}
  \caption{\textbf{Player-to-Touch comparison.} Although our hybrid model reports lower absolute accuracy/AUROC than pass-specific baselines, this is a deliberately harder and penalized setting: Player-to-Touch was trained for a different objective (first-touch hazard/receiver over time), and at inference it predicts the receiver using only hypothetical futures generated
  by the trajectory model. The objective mismatch, rollout uncertainty, and lack of training data make this a conservative comparison and likely a lower bound on performance.}
  \label{tab:touch_results}
\end{table}

\paragraph{Ball-at-Touch.}
To our knowledge, learning a module that predicts post-touch ball velocity conditioned on context and toucher identity has not been benchmarked publicly in football.
We compare to simple ablations and report likelihood-based metrics on held-out touches in Table~\ref{tab:ballatouch_results}.

\begin{table}[t]
  \centering
  \small
  \setlength{\tabcolsep}{5pt}
  \begin{tabular}{l c}
    \toprule
    Model & Masked NLL $\downarrow$ \\
    \midrule
    Constant & 0.10 \\
    Ours w/o aug. & -0.61 \\
    Ours (w/ aug.) & \textbf{-0.90} \\
    \bottomrule
  \end{tabular}
  \caption{\textbf{Ball-at-Touch ablation.} Masked negative log-likelihood (NLL) on held-out touches (lower is better).}
  \label{tab:ballatouch_results}
\end{table}

\paragraph{Possession Value (PV).}
There is extensive prior work on possession/action value, but no standardized public checkpoints or benchmarks suitable for direct comparison.
Public EPV-style implementations often depend primarily on ball location, effectively rewarding proximity to goal.
Our PV model conditions on the full player+ball configuration and predicts short-horizon goal opportunity (10\,s), which is required for scoring counterfactual rollouts in MCPS. This is shown in Table~\ref{tab:pv_results}.

\begin{table}[t]
  \centering
  \small
  \setlength{\tabcolsep}{5pt}
  \begin{tabular}{l cc}
    \toprule
    Model & Shot AUROC $\uparrow$ & Brier $\downarrow$ \\
    \midrule
    Ball-only EPV (public impl.) & \textbf{0.78} & \textbf{0.022} \\
    Ours (PV, full context) & 0.73 & 0.017 \\
    \bottomrule
  \end{tabular}
  \caption{\textbf{PV sanity checks.} Evaluation metrics for predicting shot-within-10s events. Due to lack of training data, our PV model fails to improve on discerning shot windows, even compared to a ball-only model. However, we use our model assuming that it has learned some additional context rather than distance to goal.}
  \label{tab:pv_results}
\end{table}

%% file: sec/5_discussion.tex
\section{Discussion}
\label{sec:discussion}

This paper demonstrates that modern generative multi-agent forecasting can support an MCTS-like \emph{evaluation} loop for football: a \emph{policy} that proposes counterfactual pass executions, a \emph{world model} that rolls each proposal forward under stochastic interactions, and a \emph{value model} that scores the resulting states. Instantiating these components on public tracking data yields MCPS, which evaluates an observed pass via the induced distribution over $\Delta\mathrm{PV}$ rather than a single realized outcome.

\paragraph{Takeaways.}
\textit{Distribution-aware evaluation exposes fragility.}
Local search isolates execution sensitivity by perturbing inferred kick parameters while holding the option approximately fixed; the resulting spread of $\Delta\mathrm{PV}$ (Fig.~\ref{fig:clip3}) makes ``narrow success windows'' explicit and provides a concrete notion of pass difficulty that complements point-estimate xPass-style metrics.
In practice, MCPS distinguishes passes that are valuable only under near-perfect execution from passes whose value is robust under realistic noise.

\begin{figure}
  \centering
  \includegraphics[width=0.8\linewidth]{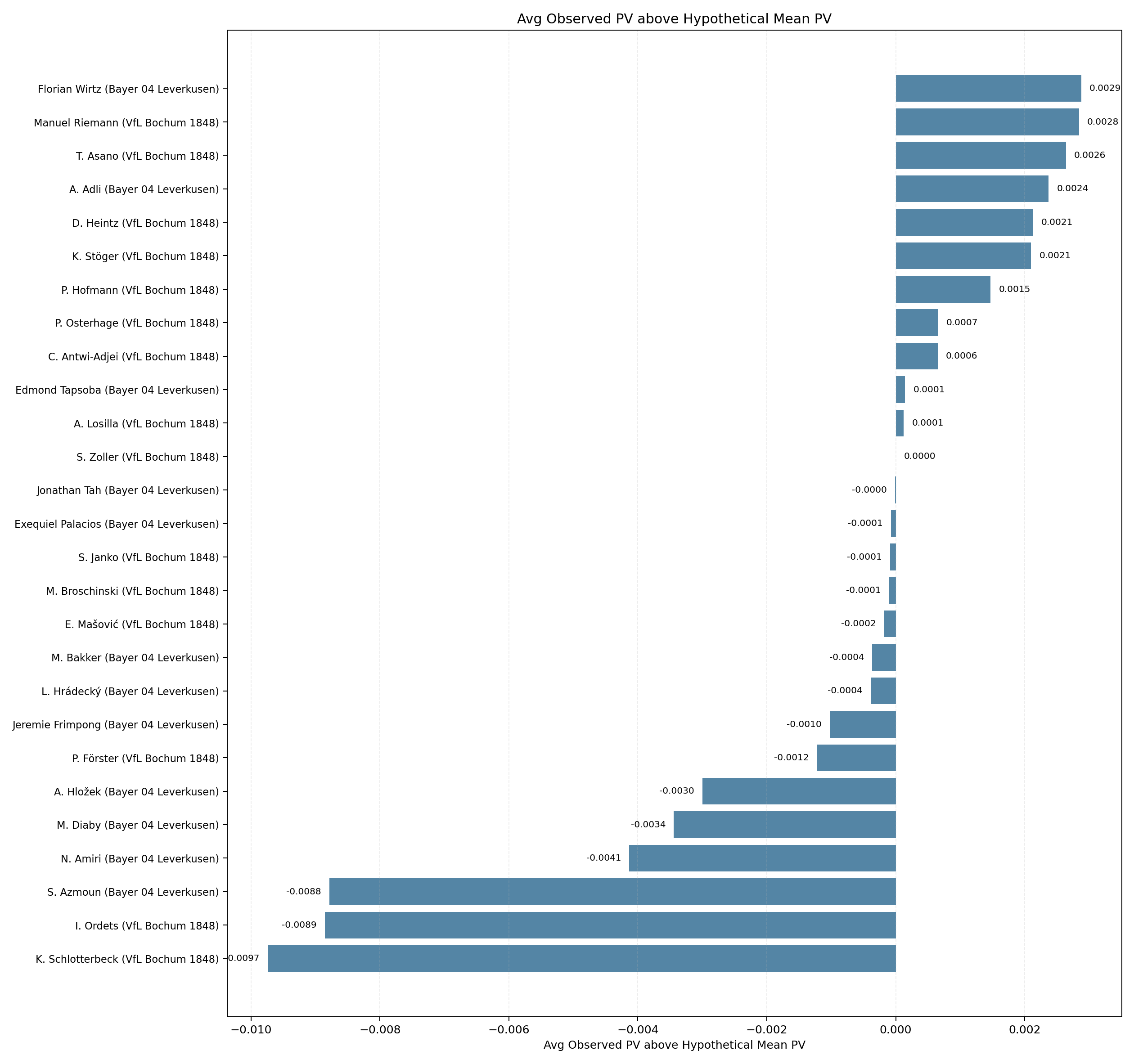}
  \caption{\textbf{Global search, mean-difference surplus.} Bars show each player's average observed-minus-counterfactual $\Delta\mathrm{PV}$ over their passes.}
  \label{fig:global}
\end{figure}

\begin{figure}
  \centering
  \includegraphics[width=0.8\linewidth]{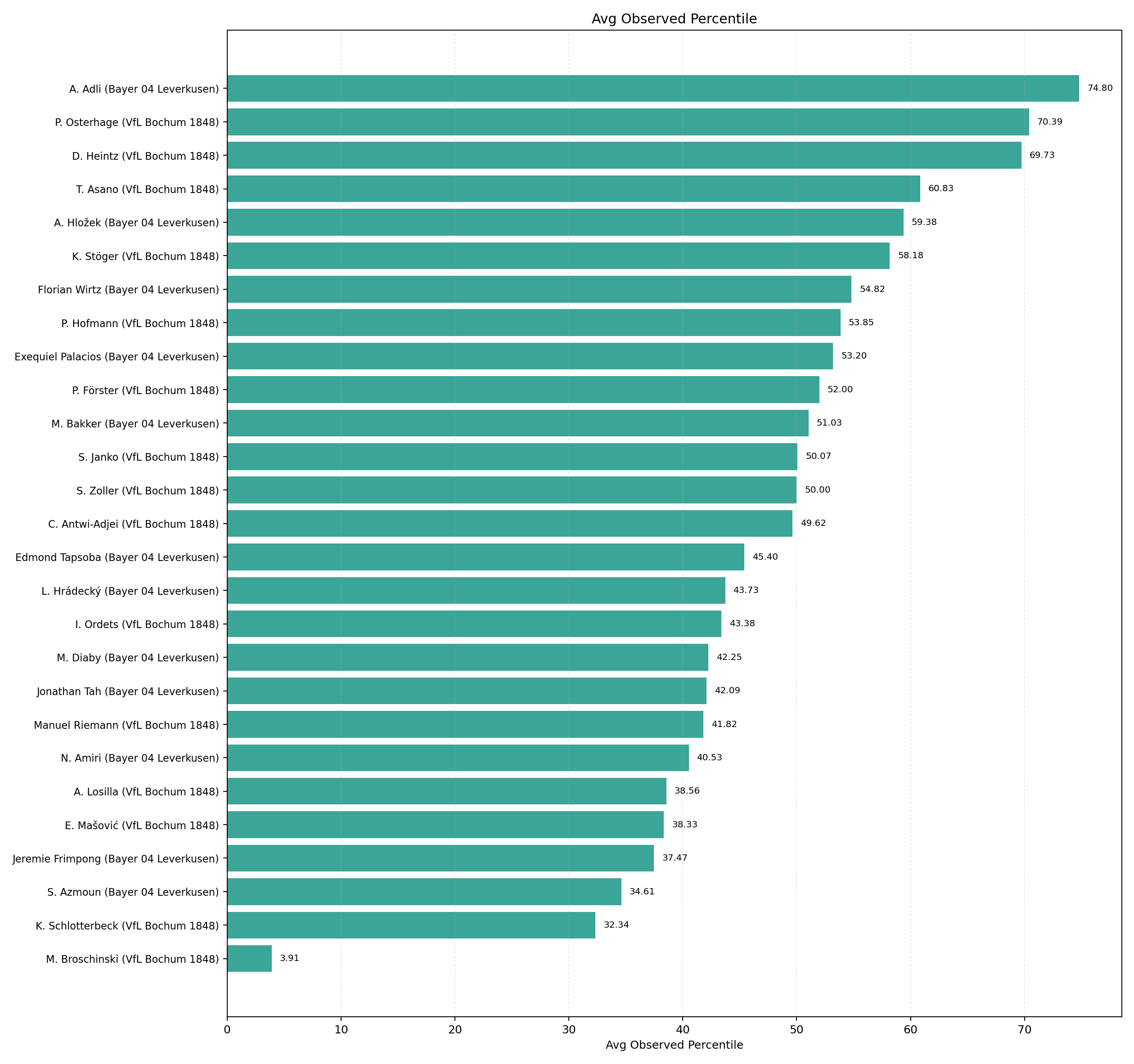}
  \caption{\textbf{Local search, percentile surplus.} Bars show each player's average percentile rank of observed-pass $\Delta\mathrm{PV}$ within the local counterfactual distribution.}
  \label{fig:local2}
\end{figure}

\textit{Local vs.\ global search disentangles execution from opportunity.}
Local variants quantify how the same intended pass might have gone under small execution changes, while global variants approximate the opportunity set available at the moment of the kick.
Aggregating these per-pass distributions yields player-level profiles that separate (i) consistent over-performance relative to execution difficulty (high percentile under local search; Fig.~\ref{fig:local2}) from (ii) systematic selection of high-value options (high observed-minus-mean under global search; Fig.~\ref{fig:global}).

\textit{Token-based rollouts are sample-efficient on scarce public data.}
Despite training on only seven matches, the discrete-token, autoregressive trajectory generator adapted from SMART provides strong best-of-$20$ forecasting accuracy (Table~\ref{tab:traj_ablation}) and supports fully hypothetical rollouts conditioned on sampled ball flights.
This suggests that language-model-style tokenization is a practical choice when public sports tracking remains small.

\paragraph{Limitations and future work.}
\textit{Data scale, noise, and synchronization.}
All sub-models are trained on a small public dataset and inherit CV tracking error (2D players, 3D ball) and event/kick misalignment.
While we mitigate this with kick-frame refinement and data augmentation, improved calibration and explicit observation-noise modeling would reduce variance and improve ranking stability.

\textit{World-model misspecification.}
Rollouts may deviate from realistic off-ball behavior, tactical conventions, and rare interaction regimes (aerial duels, goalkeeper actions, heavy contact).
Future work should incorporate player identity and role conditioning, better interaction supervision, and uncertainty estimation that separates epistemic model error from aleatoric game stochasticity.

\textit{Search distributions.}
Our local/global samplers are intentionally simple.
Global search in particular should be replaced by more realistic pass proposals (e.g., receiver-conditioned priors, pitch-control masking, or learned proposal policies) and increased diversity to better approximate the true option set, especially under tactical constraints.

\textit{Horizon and downstream value.}
We terminate rollouts at the first meaningful ball interaction for tractability and because it aligns with the immediate consequence of a pass.
Extending MCPS to multi-interaction horizons (e.g., second balls and subsequent combinations) is a natural next step, but requires careful value-model calibration to avoid compounding rollout bias.

\section*{Acknowledgements}
This material is based upon work supported by the National Science Foundation Graduate Research Fellowship Program under Grant No. DGE2140739. Any opinions,
findings, and conclusions or recommendations expressed in this material are those of the authors and do not necessarily reflect the views of the National Science Foundation.